\documentclass{article}

\usepackage[final]{tccml_neurips_2020}

\usepackage[utf8]{inputenc} 
\usepackage[T1]{fontenc}    
\usepackage{hyperref}       
\usepackage{url}            
\usepackage{booktabs}       
\usepackage{amsfonts}       
\usepackage{amsmath}
\usepackage{amssymb}
\usepackage{nicefrac}       
\usepackage{microtype}      
\usepackage[dvipsnames]{xcolor}

\usepackage{graphicx}

\title{NightVision: Generating Nighttime Satellite Imagery from Infra-Red Observations}

\author{
  Paula Harder\\
  Fraunhofer Center Machine Learning\\
  Scientific Computing, University of Kaiserlautern\\
  \texttt{paula.harder@itwm.fraunhofer.de} 
  \AND
  William Jones \\
  Dpt. of Atmospheric, Oceanic and Planetary Physics\\ University of Oxford \\
  \texttt{william.jones@physics.ox.ac.uk} 
   \And
  Redouane Lguensat \\
  LSCE-IPSL, CEA Saclay \\
  LOCEAN-IPSL, Sorbonne Université \\
  \texttt{redouane.lguensat@locean.ipsl.fr} 
   \And
  Shahine Bouabid \\
  Dpt. of Statistics\\ University of Oxford \\
  \texttt{shahine.bouabid@stats.ox.ac.uk} 
  \And
  James Fulton \\
  School of Geosciences\\ University of Edinburgh \\
  \texttt{james.fulton@ed.ac.uk} 
  \And
  Dánnell Quesada-Chacón \\
  Institute of Hydrology and Meteorology\\ Dresden University of Technology \\
  \texttt{dannell.quesada@tu-dresden.de} 
  \And
  Aris Marcolongo \\
  Mathematical Institute\\ University of Bern \\
  Climate and Environmental Physics\\ University of Bern \\
  \texttt{aris.marcolongo@math.unibe.ch} 
  \And
  Sofija Stefanović \\
  Dpt. of Atmospheric, Oceanic and Planetary Physics\\ University of Oxford\\
  \texttt{sofija.stefanovic@physics.ox.ac.uk} 
  \And
  Yuhan Rao \\
  Institute for Climate Studies\\ North Carolina State University \\
  \texttt{yrao5@ncsu.edu} 
  \And
  Peter Manshausen \\
  Dpt. of Atmospheric, Oceanic and Planetary Physics\\ University of Oxford\\
  \texttt{peter.manshausen@physics.ox.ac.uk} 
  \And
  Duncan Watson-Parris \\
  Dpt. of Atmospheric, Oceanic and Planetary Physics\\ University of Oxford \\
   \texttt{duncan.watson-parris@physics.ox.ac.uk} 
}

\begin{document}

\maketitle

\begin{abstract}
The recent explosion in applications of machine learning to satellite imagery often rely on visible images and therefore suffer from a lack of data during the night. The gap can be filled by employing available infra-red observations to generate visible images. This work presents how deep learning can be applied successfully to create those images by using U-Net based architectures. The proposed methods show promising results, achieving a structural similarity index (SSIM) up to 86\% on an independent test set and providing visually convincing output images, generated from infra-red observations. The code is available at: \url{https://github.com/paulaharder/hackathon-ci-2020}
\end{abstract}

\section{Introduction}

The availability of huge amounts of open, high-quality satellite imagery from e.g. Sentinel-2 and GOES-16 has enabled many machine learning (ML) applications as diverse as the detection of Penguins \citep{doi:10.1002/rse2.176}, solar-panels~\citep{hou2019solarnet} and ship-tracks \citep{cloud_pertubations}. Many of these applications rely on visible imagery because of the prevalence of pre-existing models and easier processing and validation. This visible imagery relies on the detection of reflected sunlight during the day, whereas instruments such as the Advanced Baseline Imager (ABI) on-board GOES-16 are also able to measure infra-red emission throughout the day and night. These different spectra contain different, but often complementary information, and in principle ML models could be trained to use either or both depending on their availability. In practice, having access to homogenised imagery allows a single ML model to be trained to, for example, detect and track clouds 24 hours a day. This capability would transform our ability to detect the subtle, but important, perturbations humans are exerting on the climate system \citep{Stevens2009}.

Here we present the result of a three-day hackathon which challenged contestants to generate visible (RGB) images using only the infra-red imagery available at night. We describe the challenge and introduce the publicly available training datasets in Section 2, present the three winning models and their notable features in Section 3, and discuss avenues of future work in Section 4. 

We are not aware of any work on generating visible satellite imagery from infra-red observation. In
\citep{berg2018generating} thermal infra-red observations are used to generate visible spectrum images for traffic scene datasets, also by employing convolutional neural networks.

\section{Data Preparation}
Data is acquired by the ABI aboard the Geostationary Operational Environment Satellite (GOES)-16 \citep{schmit_goes-r_2008}. This is a modern Earth Observation (EO) platform placed in a geostationary orbit, allowing it to provide visible and IR imagery every ten minutes. Channels 8-16 of infra-red (IR) channels from the GOES-16 ABI instrument are used to create the inputs of our algorithm, while channels 1-6, which detect reflected solar radiation, are used to create the visible target outputs. Additional information about the channels used by ABI, and what physical properties they measure, can be found here\footnote{\url{https://www.goes-r.gov/mission/ABI-bands-quick-info.html}}.

For ease of processing we transform these raw channel radiances into RGB composite images using SatPy software \citep{martin_raspaud_2020_4036291} for both the IR channels (model input) and visible (target outut). The considered region is -85 to -70 degrees longitude and -15 to -30 degrees latitude, and to reduce the size of the data we downsample the images to a size of $127\times127$ pixels.

For the competition, 2 years of data are provided as training dataset from January 1\textsuperscript{st} 2018 to December 15\textsuperscript{th} 2019, the two last weeks of December are thrown out to avoid data leakage between train and test datasets. The competing methods were tuned first in a validation phase where data from the first 15 days of January 2020 were used to calculate the metrics. In the test phase, the participants used their best performing method once on a test dataset consisting of the last 15 days of January 2020, those results are the ones reported in this paper and served for the final ranking.

Interested readers have the possibility of participating in a public version of the challenge on the Codalab platform\footnote{\url{https://competitions.codalab.org/competitions/26644}}. Data related to the challenge is available from Zenodo repositories (\cite{redouane_lguensat_2020_4061336}).

\section{Generating Nighttime Imagery with Deep Learning}

\subsection{Methodology}

To tackle the problem of generating visible light images during the night, we employ three different approaches, supervised Conditional GANs (cGANs), a U-Net and a U-Net ++. All the methods are U-Net based and share the use of the structural similarity index methods (SSIM)~\citep{SSIM} in their (generator) loss function:
$$\text{SSIM}(x,y)=\frac{(2\mu_x\mu_y+c_1)(2\sigma_{xy}+c_2)}{(\mu_x^2+\mu_y^2+c_1)(\sigma_x^2+\sigma_y^2+c_2)}, $$
where $x,y$ are two images, $\mu_x,\mu_y$ their mean values, $\sigma_x,\sigma_y$ their variances, $\sigma_{xy}$ their covariance, all calculated over all pixels and channels. The variables $c_1, c_2$ are  added to stabilize the division with weak denominator\footnote{here $c_1=(0.01\cdot L)^2$, $c_2=(0.03\cdot L)^2$, $L$ is the dynamic range of the pixel values}.
As the models were developed independently from each other during a coding competition, they also show different data preprocessing and different experimental setups.

\subsubsection{Data Preprocessing}
\paragraph{Black Images/Pixels}The underlying data set contains about 8,000 pairs of infra-red images and visible light images, both from the night and the day. For the training only daylight images are of use, therefore we sort out nighttime images. Images from around sunset/sunrise, which contain many black pixels, are treated differently by the three approaches. The data preprocessing for the cGANs method leaves most of the sunrise/sunset pictures in the training set, by only sorting out those with more than 99\% black pixels, whereas the U-Net approach only includes images in the training set with a percentage of non-zero entries across the three RGB channels higher than 80\%. For the U-Net ++ based method we use a more complex approach. We filter the initial set of images to remove those where more than 0.2\% of pixels are `dark' in the visible image channels. We label a pixel as `dark' if the sum of pixel values across the 3 visible channels is less than 5 (out of a maximum of 765). The remaining dark pixels in the filtered data are assigned a NaN value which is used to mask them when calculating SSIM. From our observations these dark pixels mainly occur in parts of the image around sunrise and sunset, where the sun is not shining on a corner of the image. Only a very small proportion of dark pixels comes from images which are fully in daylight.
We initially had a much more lenient filter for dark images based on minimum average pixel value. However this permitted training images from around sunrise or sunset where a significant portion of the image was in darkness. This caused the trained network to predict dark patches as well, which is counter to our objective. 


\subsubsection{Models}

\paragraph{Baseline Model} As a baseline model we use the k-nearest neighbour (kNN) regression, with $k=3$. The kNN regression means for every infra-red image in the test set we look at the three closest infra-red images in the training set and then predict the mean of their visible counter parts.

\paragraph{U-Net/U-Net ++} Based on the fully convolutional network \citep{fully_conv} the U-Net \citep{unet} was developed for biological image segmentation. The network consists of similar downsampling and upsampling parts, which yields a u-shape architecture. The U-Net skip connections also allow the spatial structure of the input thermal image to shuttle across the net. The U-Net++ \citep{zhou2018unet++} model is a recent and more powerful iteration on the U-Net architecture with enhanced skip connections. Similar to the U-Net, this model allows the final layer of the network to utilise fine grain detail from shallow paths through the network, and coarse grain context from the downsampled then upsampled paths. This architecture has been highly successful for image segmentation, but has been used little for image-to-image regression as in this task. We implement the U-Net++ both with and without deep supervision \citep{lee2015deeply}. We find that the model with deep supervision takes longer to converge and makes less accurate predictions on the test set, particularly on images with large spatially homogeneous cloud fields.

\paragraph{Loss Functions}
Both U-Net approaches uses a SSIM based loss function. For the U-Net ++ we implement a slightly modified version of SSIM so that dark pixels from the target image are ignored when the spatial average of SSIM is taken. As we use a window size of 11, this means that a single dark pixel in the target image masks an area within a 5 pixel radius of itself. No masking is applied if the predicted pixel is dark.

\paragraph{Supervised cGANs}

cGANs are a class of generative models where a generator $G$ learns a mapping from a random noise $z$ and a conditioning input $x$, to an output $y = G(z|x)$. The generator grows by attempting to fool a discriminator $D$ who learns itself to estimate the likelihood $D(y|x)$ of sample $y$ being either real or generated by $G$. We propose to frame the nighttime visible imaging from thermal infra-red observation in the cGANs rationale. Namely, given an infra-red sample $x$ we train a generator to estimate its corresponding optical image as $\hat y = G(z|x)$. The discriminator enables the generator to capture and reproduce important realistic features. We augment the objective with a supervised loss to foster generation of images close to the ground truth. Our preference goes for an $L_1$ penalty to capture low-frequency components while inducing less blurring than the $L_2$ norm. In addition, we also use a structural similarity index based (SSIM)~\citep{SSIM} supervision to encourage perceptual persistence of pixels, captured by high-frequency components. The additional discriminator enables the model to find important features to measure similarity other than the closeness in a $L_1$ and SSIM sense. Eventually, the infrared-to-optical image translation problem writes as the two-player game
\begin{equation}
\min_G\max_D \,\mathcal{L}_{\text{cGAN}}(G, D) + \lambda\,\mathcal{L}_{L_1}(G) + \mu\,\mathcal{L}_{\operatorname{SSIM}}(G)
\end{equation}
where $\mathcal{L}_{\text{cGAN}}(G, D) = \mathbb{E}\left[\log D(y|x)\right] + \mathbb{E}\left[\log\left(1 - D(G(z|x)|x)\right)\right]$, $\mathcal{L}_{L_1} = \mathbb{E}\left[\|G(z|x) - y\|_1\right]$ and $\mathcal{L}_{\operatorname{SSIM}}(G) = \mathbb{E}\left[1 - \operatorname{SSIM}(G(z|x), y)\right]$. Expectations are taken over all couples of infra-red and optical images $(x, y)$. 

Inspired by the success of pix2pix~\citep{isola17} based approaches in remote sensing, we use a U-Net architecture for the generator and a PatchGAN discriminator~\citep{isola17}.  Instead of feeding the generator with random noise, we provide stochasticity with dropout layers in the decoder. Unlike usual discriminators, PatchGAN classifies jointly local regions of the image, fostering generation of high-frequency components on top of the SSIM supervision.

\subsubsection{Experimental Setups}

The network in the pure U-Net approach consists of 6 levels of depth, in the other methods we use U-Nets with 5 levels each. All methods use an Adam optimizer \citep{Adam}. For the cGAN model, to make up for the additional guidance provided to the generator by the supervision objectives, we backpropagate on the discriminator twice as much. Using $\lambda$ = 0.01, $\mu$ = 10, it is trained with an initial learning rate 2e-4, decay 0.99, for 500 epochs.  We train the U-Net with constant rate 1e-3 and early stopping for 54 epochs. The U-Net++ model is trained at a rate of 2e-1 for (60, 30, 30, 20, 20) epochs using batch sizes of (10, 32, 64, 128, 256) respectively so that gradient updates in later epochs are subject to less noise.

\subsection{Results}

The three proposed models are tested on an independent set of 200 test images from daylight (since ground truth is available), showing the same scenery as the training images. In Table \ref{Tab:scores} we report the SSIM and the root mean squared error (RSME). All method show a huge improvement compared to the baseline method. The cGAN approach reaches a 0.77 mean SSIM, whereas the U-Net method is about 10\% higher. The U-Net++ has slightly better scores, both mean SSIM and RSME, than its simpler version. 
	\begin{table}[h]
		\caption{Metrics for the different methods}
		\label{Tab:scores}
		\centering
		\begin{tabular}{lll}
			Method     & SSIM     & RSME \\
			\midrule
			Baseline: kNN & 0.15 &0.24\\
			Method 1: cGAN     & 0.77       & 0.11  \\
			Method 2: U-Net     & 0.85 & 0.09      \\
			Method 3: U-Net++ & 0.86  & 0.07    \\
			\bottomrule
		\end{tabular}
	\end{table}

\begin{figure}[h]
    \centering
    \includegraphics[width=\textwidth]{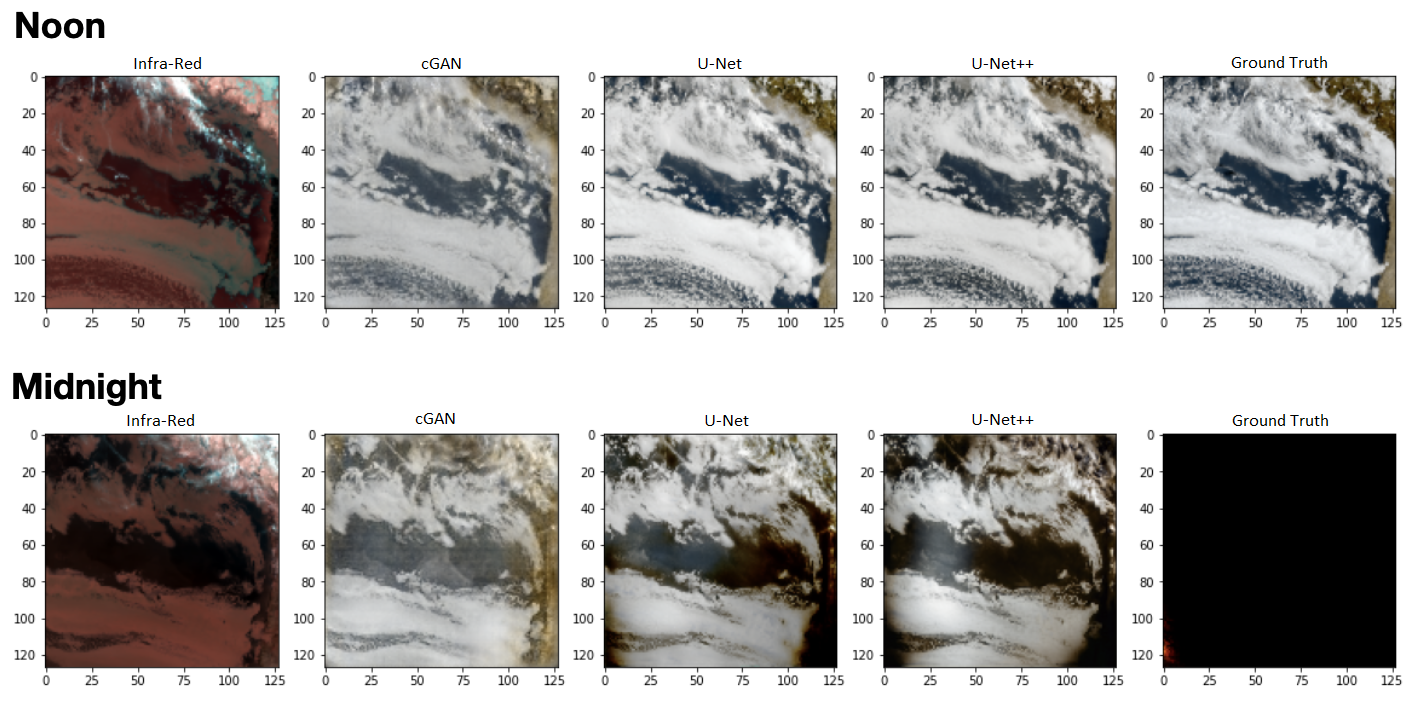}
    \caption{An example from the application of the three methods on unseen images during training, two case are shown for January 29\textsuperscript{th} 2020: at midnight and at noon.}
    \label{fig:my_label}
\end{figure}

As the human eye is probably the best measure for the quality of the generated images we look at an example.
The first row of Figure \ref{fig:my_label} shows a randomly picked test image during the day. All methods are able to correctly color clouds, sea and land, subtle details from the clouds are reproduced. The image we generated with the U-Net++ model is hardly distinguishable from the ground truth. With our cGAN model we also manage to capture the shapes and details, but the colors are slightly bleached out.
As the actual goal is to predict images from infra-red observations during the night, row two of Figure \ref{fig:my_label} shows the results of the models applied on a midnight image. The shapes of the clouds are reproduced nicely by each model. The U-Net/U-Net++ generated images show black areas on the edges and color some parts of the ocean darkly, in the U-Net++ approach this is more strongly pronounced. The images generated by the cGAN method shows good coloring, but some artifacts from the GAN patches show up and we can see horizontal and vertical lines.

\section{Conclusion}
In this work we showed that different U-Net based models are capable of producing a visible spectrum image from IR observations. Especially the non-GAN approaches show very high quality synthetic images for the daytime observations, on the other hand for the final goal, generating from nighttime observations, the GAN approach images show some desired properties better than the other two methods. This paper presented first experiments, showing promising results. Further work needs to be done to deal with black spots in nighttime predictions and provide a consistent experimental setup to explore the proposed methods in detail and have more comparable results. Future research should aim to generalize model performance better from day to night, taking the subtle differences of IR observations between night and day into account.

\subsection{Acknowledgments}

We would like to thank the organizers of the Climate Informatics Conference, especially the Hackathon chair as well as all participants. We are grateful for the reviewers helpful comments.
DWP receives funding from the European Union’s Horizon 2020 research and innovation programme iMIRACLI under Marie Skłodowska-Curie grant agreement No 860100 and also gratefully acknowledges funding from the NERC ACRUISE project NE/S005390/1. AM acknowledges funding from the Swiss National Science Foundation (grant number 189908).

\bibliography{imagery}

\end{document}